\documentclass[11pt]{article}

\usepackage{colacl,amsmath,rotate,psfig,graphicx,epsf}

\author{Franz Beil, Glenn Carroll, Detlef Prescher, Stefan Riezler
  \and Mats Rooth\\ Institut f\"ur Maschinelle Sprachverarbeitung,
  University of Stuttgart}

\title{Inside-Outside Estimation of a Lexicalized PCFG for German}

\begin{document}

\bibliographystyle{acl}

\maketitle

\begin{abstract}
  The paper describes an extensive experiment in inside-outside
  estimation of a lexicalized probabilistic context free grammar for
  German verb-final clauses. Grammar and formalism features which make
  the experiment feasible are described.  Successive models are
  evaluated on precision and recall of phrase markup.
\end{abstract}

\section{Introduction}

\newcite{Charniak:95} and \newcite{Carroll/Rooth:98} present 
head-lexicalized probabilistic context free grammar formalisms,
and show that they can effectively be applied in inside-outside
estimation of syntactic language models for English, the
parameterization of which encodes lexicalized rule probabilities and
syntactically conditioned word-word bigram collocates.  The present
paper describes an experiment where a slightly modified version of
Carroll and Rooth's model was applied in a systematic experiment on
German, which is a language with rich inflectional morphology and free
word order (or rather, compared to English, free-er phrase order). We
emphasize techniques which made it practical to apply inside-outside
estimation of a lexicalized context free grammar to such a language.
These techniques relate to the treatment of argument cancellation and
scrambled phrase order; to the treatment of case features in category
labels; to the category vocabulary for nouns, articles, adjectives and
their projections; to lexicalization based on uninflected lemmata
rather than word forms; and to exploitation of a parameter-tying
feature.

\section{Corpus and morphology}
\label{sec:corpus}

The data for the experiment is a corpus of German subordinate clauses
extracted by regular expression matching from a 200 million token
newspaper corpus. The clause length ranges between four and 12 words.
Apart from infinitival VPs as verbal arguments, there are no further
clausal embeddings, and the clauses do not contain any punctuation
except for a terminal period. The corpus contains 4128873 tokens and
450526 clauses which yields an average of 9.16456 tokens per clause.
Tokens are automatically annotated with a list of part-of-speech (PoS)
tags using a computational morphological analyser based on
finite-state technology (\newcite{Karttunen/Yampol/Grefenstette:94},
\newcite{Schiller/Stoeckert:95}).

A problem for practical inside-outside estimation of an inflectional
language like German arises with the large number of terminal and
low-level non-terminal categories in the grammar resulting from the
morpho-syntactic features of words. Apart from major class (noun,
adjective, and so forth) the analyser provides an ambiguous word with
a list of possible combinations of inflectional features like gender,
person, number (cf.\ the top part of Fig.\ \ref{fig:infl} for an
example ambiguous between nominal and adjectival PoS; the PoS is
indicated following the '\texttt{+}' sign). In order to reduce the
number of parameters to be estimated, and to reduce the size of the
parse forest used in inside-outside estimation, we collapsed the
inflectional readings of adjectives, adjective derived nouns, article
words, and pronouns to a single morphological feature (see of Fig.\ 
\ref{fig:infl} for an example).  This reduced the number of low-level
categories, as exemplified in Fig.\ \ref{fig:corpus}: {\em das} has
one reading as an article and one as a demonstrative; {\em
  westdeutschen} has one reading as an adjective, with its
morphological feature N indicating the inflectional suffix.

\begin{figure}[h]
  \begin{center}\small
\begin{verbatim}
  analyze> Deutsche
  1. deutsch^ADJ.Pos+NN.Fem.Akk.Sg
  2. deutsch^ADJ.Pos+NN.Fem.Nom.Sg
  3. deutsch^ADJ.Pos+NN.Masc.Nom.Sg.Sw
  4. deutsch^ADJ.Pos+NN.Neut.Akk.Sg.Sw
  5. deutsch^ADJ.Pos+NN.Neut.Nom.Sg.Sw
  6. deutsch^ADJ.Pos+NN.NoGend.Akk.Pl.St
  7. deutsch^ADJ.Pos+NN.NoGend.Nom.Pl.St
  8. *deutsch+ADJ.Pos.Fem.Akk.Sg
  9. *deutsch+ADJ.Pos.Fem.Nom.Sg
  10. *deutsch+ADJ.Pos.Masc.Nom.Sg.Sw
  11. *deutsch+ADJ.Pos.Neut.Akk.Sg.Sw
  12. *deutsch+ADJ.Pos.Neut.Nom.Sg.Sw
  13. *deutsch+ADJ.Pos.NoGend.Akk.Pl.St
  14. *deutsch+ADJ.Pos.NoGend.Nom.Pl.St
\end{verbatim}
\begin{verbatim}
  ==>   Deutsche  { ADJ.E, NNADJ.E }
\end{verbatim}
    \vspace{-1em}
    \caption{Collapsing Inflectional Features}
    \label{fig:infl}
  \end{center}
\end{figure}



We use the special tag \texttt{UNTAGGED} indicating that the analyser
fails to provide a tag for the word. The vast majority of
\texttt{UNTAGGED} words are proper names not recognized as such. These
gaps in the morphology have little effect on our experiment.

\begin{figure}[ht]
  \begin{center}\small{}
\begin{verbatim}
  während   { ADJ.Adv, ADJ.Pred, KOUS, 
              APPR.Dat, APPR.Gen }
  sich  { PRF.Z }
  das   { DEMS.Z, ART.Def.Z }
  Preisniveau   { NN.Neut.NotGen.Sg }
  dem   { DEMS.M, ART.Def.M }
  westdeutschen     { ADJ.N }
  annähere  { VVFIN }
  . { PER }
\end{verbatim}
    \caption{Corpus Clip}
    \label{fig:corpus}
  \end{center}
  \vspace{-1em}
\end{figure}



\section{Grammar}
\label{sec:grammar}

The grammar is a manually developed headed context-free phrase
structure grammar for German subordinate clauses with 5508 rules
and 562 categories, 209 of which are terminal categories. The
formalism is that of \newcite{Carroll/Rooth:98}, henceforth C+R:
{\small
\begin{verbatim}
 mother -> non-heads head' non-heads (freq)
\end{verbatim}
}

The rules are head marked with a prime. The non-head sequences may be
empty. \texttt{freq} is a rule frequency, which is initialized
randomly and subsequently estimated by the inside outside-algorithm.
To handle systematic patterns related to features, rules were
generated by Lisp functions, rather than being written directly in the
above form. With very few exceptions (rules for coordination, S-rule),
the rules do not have more than two daughters.

\begin{figure*}[ht]
  \begin{center}
    \mbox{\psfig{file=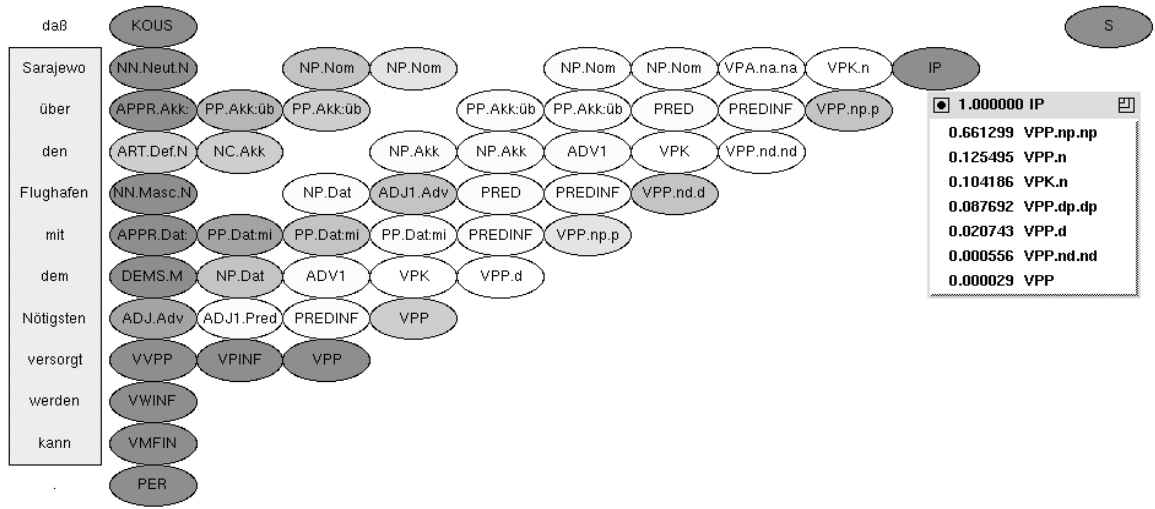,angle=270,width=40em,height=18em}}
    \caption{Chart browser}
    {\small Word-by-word gloss of the clause:\textsl{'that
      Sarajevo over the airport with the essentials supplied will
      can'}}\label{fig:chart}
\end{center}
\vspace{-1em}
\end{figure*}

Grammar development is facilitated by a chart browser that permits a
quick and efficient discovery of grammar bugs \cite{Carroll:97b}.
Fig.\ \ref{fig:chart} shows that the ambiguity in the chart is quite
considerable even though grammar and corpus are restricted. For the
entire corpus, we computed an average 9202 trees per clause. In the
chart browser, the categories filling the cells indicate the most
probable category for that span with their estimated frequencies. The
pop-up window under \texttt{IP} presents the ranked list of all
possible categories for the covered span. Rules (chart edges) with
frequencies can be viewed with a further menu. In the chart browser,
colors are used to display frequencies (between 0 and 1) estimated by
the inside-outside algorithm. This allows properties shared across
tree analyses to be checked at a glance; often grammar and estimation
bugs can be detected without mouse operations.

The grammar covers 88.5\% of the clauses and 87.9\% of the tokens
contained in the corpus. Parsing failures are mainly due to
\texttt{UNTAGGED} words contained in 6.6\% of the failed clauses,
the pollution of the corpus by infinitival constructions
($\approx$1.3\%), and a number of coordinations not covered by the
grammar ($\approx$1.6\%).

\subsection{Case features and agreement}

On nominal categories, in addition to the four cases Nom, Gen, Dat,
and Akk, case features with a disjunctive interpretation (such as Dir
for Nom or Akk) are used.  The grammar is written in such a way that
non-disjunctive features are introduced high up in the tree. This
results in some reduction in the size of the parse forest, and some
parameter pooling. Essentially the full range of agreement inside the
noun phrase is enforced.  Agreement between the nominative NP and the
tensed verb (e.g. in number) is not enforced by the grammar, in order
to control the number of parameters and rules.

For noun phrases we employ Abney's chunk grammar organization
\cite{Abney:96}. The noun chunk (NC) is an approximately non-recursive
projection that excludes post-head complements and (adverbial)
adjuncts introduced higher than pre-head modifiers and determiners but
includes participial pre-modifiers with their complements. Since we
perform complete context free parsing, parse forest construction, and
inside-outside estimation, chunks are not motivated by deterministic
parsing. Rather, they facilitate evaluation and graphical debugging,
by tending to increase the span of constituents with high estimated
frequency.

\subsection{Subcategorisation frames of verbs}

The grammar distinguishes four subcategorisation frame classes: active
(VPA), passive (VPP), infinitival (VPI) frames, and copula
constructions (VPK). A frame may have maximally three arguments.
Possible arguments in the frames are nominative (n), dative (d) and
accusative (a) NPs, reflexive pronouns (r), PPs (p), and infinitival
VPs (i). The grammar does not distinguish plain infinitival VPs from
{\em zu}-infinitival VPs. The grammar is designed to partially
distinguish different PP frames relative to the prepositional head of
the PP.  A distinct category for the specific preposition becomes
visible only when a subcategorized preposition is cancelled from the
subcat list. This means that specific prepositions do not figure in
the evaluation discussed below. The number and the types of frames in
the different frame classes are given in figure \ref{fig:frames}.

\begin{figure}[t]
\small
  \begin{center}
    \begin{tabular}{lcl}
      class & \# & frame types\\\hline
      VPA &
      15 &
      n, na, nad, nai, nap, nar, nd, ndi,\\ && ndp, ndr, ni, nir, np, npr, nr\\
      VPP &
      13 &
      d, di, dp, dr, i, ir, n, nd, ni, np, p,\\ &&pr, r \\
      VPI &
      10 &
      a, ad, ap, ar, d, dp, dr, p, pr, r \\
      VPK &
      2 &
      i, n 
    \end{tabular}
    \caption{Number and types of verb frames}
    \label{fig:frames}
  \end{center}
  \vspace{-1em}
\end{figure}



\begin{figure}[t]
  \begin{center}
    \epsfbox{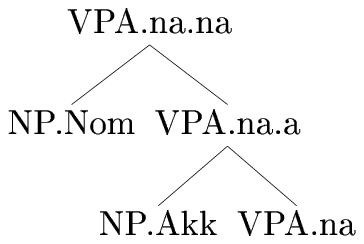}\quad
    \epsfbox{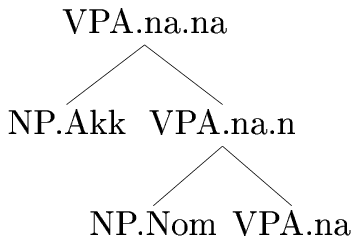}
    \caption{Coding of canonical and scrambled argument order}
    \label{fig:vp-trees}
  \end{center}
  \vspace{-1em}
\end{figure}

German, being a language with comparatively free phrase order, allows
for scrambling of arguments. Scrambling is reflected in the particular
sequence in which the arguments of the verb frame are saturated.
Compare figure \ref{fig:vp-trees} for an example of a canonical
subject-object order in an active transitive frame and its scrambled
object-subject order. The possibility of scrambling verb arguments
yields a substantial increase in the number of rules in the grammar
(e.g.\ 102 combinatorically possible argument rules for all in VPA
frames). Adverbs and non-subcategorized PPs are introduced as adjuncts
to VP categories which do not saturate positions in the subcat frame.

In earlier experiments, we employed a flat clausal structure, with
rules for all permutations of complements.  As the number of frames
increased, this produced prohibitively many rules, particularly with
the inclusion of adjuncts.

\section{Parameters}

The parameterization is as in C+R, with one significant modification.
Parameters consist of (i) rule parameters, corresponding to right hand
sides conditioned by parent category and parent head; (ii) lexical
choice parameters for non-head children, corresponding to child lemma
conditioned by child category, parent category, and parent head lemma.
See C+R or \newcite{Charniak:95} for an explanation of how such
parameters define a probabilistic weighting of trees.  The change
relative to C+R is that lexicalization is by uninflected lemma rather
than word form. This reduces the number of lexical parameters, giving
more acceptable model sizes and eliminating splitting of estimated
frequencies among inflectional forms.  Inflected forms are generated
at the leaves of the tree, conditioned on terminal category and lemma.
This results in a third family of parameters, though usually the
choice of inflected form is deterministic.

A parameter pooling feature is used for argument filling where all
parent categories of the form VP.x.y are mapped to a category VP.x in
defining lexical choice parameters.  The consequence is e.g.\ that an
accusative daughter of a nominative-accusative verb uses the same
lexical choice parameter, whether a default or scrambled word order is
used.  (This feature was used by C+R for their phrase trigram grammar,
not in the linguistic part of their grammar.) Not all desirable
parameter pooling can be expressed in this way, though; for instance
rule parameters are not pooled, and so get split when the parent
category bears an inflectional feature.

\section{Estimation}
\label{sec:estimation}

The training of our probabilistic CFG proceeds in three steps: (i)
unlexicalized training with the \texttt{supar} parser, (ii)
bootstrapping a lexicalized model from the trained unlexicalized one
with the \texttt{ultra} parser, and finally (iii) lexicalized training
with the \texttt{hypar} parser \cite{Carroll:97a}. Each of the three
parsers uses the inside-outside algorithm. \texttt{supar} and
\texttt{ultra} use an unlexicalized weighting of trees, while
\texttt{hypar} uses a lexicalized weighting of trees. \texttt{ultra}
and \texttt{hypar} both collect frequencies for lexicalized rule and
lexical choice events, while \texttt{supar} collects only
unlexicalized rule frequencies.

Our experiments have shown that training an unlexicalized model first
is worth the effort. Despite our use of a manually developed grammar
that does not have to be pruned of superfluous rules like an
automatically generated grammar, the lexicalized model is notably
better when preceded by unlexicalized training (see also
\newcite{Ersan/Charniak:95} for related observations). A comparison of
immediate lexicalized training (without prior training of an
unlexicalized model) and our standard training regime that involves
preliminary unlexicalized training speaks in favor of our strategy
(cf.\ the different 'lex 0' and 'lex 2' curves in figures
\ref{fig:nc-plot} and \ref{fig:vp-plot}). However, the amount of
unlexicalized training has to be controlled in some way.

A standard criterion to measure overtraining is to compare
log-likelihood values on held-out data of subsequent iterations. While
the log-likelihood value of the training data is theoretically
guaranteed to converge through subsequent iterations, a decreasing
log-likelihood value of the held-out data indicates overtraining.
Instead of log-likelihood, we use the inversely proportional
cross-entropy measure. Fig.~\ref{fig:overtraining} shows comparisons
of different sizes of training and heldout data (training/heldout):
(A) 50k/50k, (B) 500k/500k, (C) 4.1M/500k. The overtraining effect is
indicated by the increase in cross-entropy from the penultimate to the
ultimate iteration in the tables. Overtraining results for lexicalized
models are not yet available.

\begin{figure}[t]
\footnotesize
\begin{center}
\begin{tabular}[t]{r@{:\quad}r@{.}l}
\multicolumn{3}{c}{{\bf A}}\\\hline
1 & 52&0199\\
2 & 25&3652\\
3 & 24&5905\\
\multicolumn{1}{c}{\vdots} & \multicolumn{1}{c}{\vdots}\\
13 & 24&2872\\
14 & 24&2863\\
15 & 24&2861\\
16 & 24&2861\\
17 & 24&2867
\end{tabular}\hspace{1em}
\begin{tabular}[t]{r@{:\quad}r@{.}l}
\multicolumn{3}{c}{{\bf B}}\\\hline
1 & 53&7654\\
2 & 26&3184\\
3 & 25&5035\\
\multicolumn{1}{c}{\vdots} & \multicolumn{1}{c}{\vdots}\\
55 & 25&0548\\
56 & 25&0549\\
57 & 25&0549\\
58 & 25&0549\\
59 & 25&055
\end{tabular}\hspace{1em}
\begin{tabular}[t]{r@{:\quad}r@{.}l}
\multicolumn{3}{c}{{\bf C}}\\\hline
1 & 49&8165\\
2 & 23&1008\\
3 & 22&4479\\
\multicolumn{1}{c}{\vdots} & \multicolumn{1}{c}{\vdots}\\
70 & 22&1445\\
80 & 22&1443\\
90 & 22&1443\\
95 & 22&1443\\
96 & 22&1444
\end{tabular}
    \caption{\label{fig:overtraining}Overtraining (iteration:
      cross-entropy on heldout data)}
  \end{center}
  \vspace{-2em}
\end{figure}



\begin{figure*}[t]
\begin{center}
    \mbox{\psfig{file=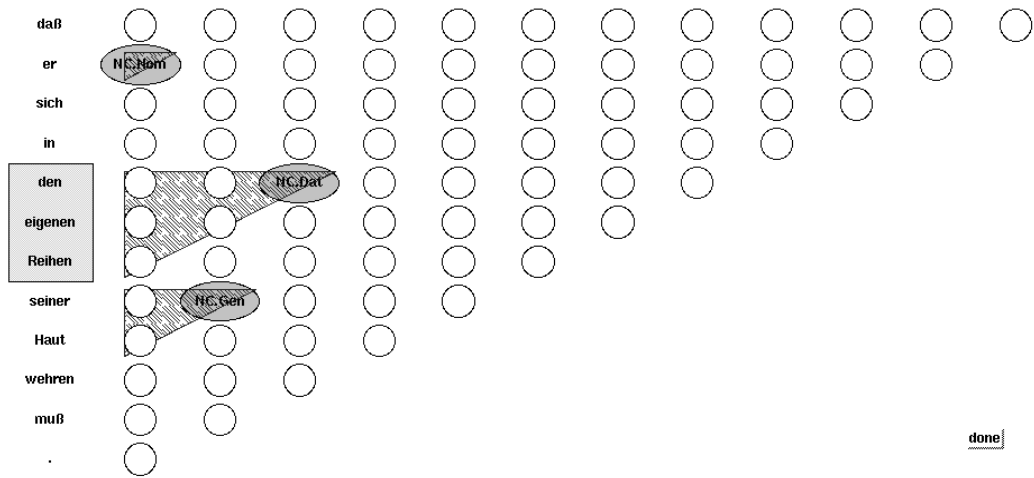,angle=270,width=36em,height=17em}}
    \vspace{-1em}
    \caption{Chart browser for manual NC labelling}
    \label{fig:muon}
  \end{center}
  \vspace{-1em}
\end{figure*}

However, a comparison of precision/recall measures on categories of
different complexity through iterative unlexicalized training shows
that the mathematical criterion for overtraining may lead to bad
results from a linguistic point of view. While we observed more or
less converging precision/recall measures for lower level structures
such as noun chunks, iterative unlexicalized training up to the
overtraining threshold turned out to be disastrous for the evaluation
of complex categories that depend on almost the entire span of the
clause. The recognition of subcategorization frames through 60
iterations of unlexicalized training shows a massive decrease in
precision/recall from the best to the last iteration, even dropping
below the results with the randomly initialized grammar (see Fig.\ 
\ref{fig:vp-plot}).

\subsection{Training regime}
\label{sec:training}

We compared lexicalized training with respect to different starting
points: a random unlexicalized model, the trained unlexicalized model
with the best precision/recall results, and an unlexicalized model that
comes close to the cross-entropy overtraining threshold. The details
of the training steps are as follows:
\begin{enumerate}\renewcommand{\labelenumi}{(\arabic{enumi})\ }
  \setlength{\itemsep}{-0.8ex}
\item 0, 2 and 60 iterations of unlexicalized parsing with
  \texttt{supar};
\item lexicalization with \texttt{ultra} using the entire corpus;
\item 23 iterations of lexicalized parsing with \texttt{hypar}.
\end{enumerate}

The training was done on four machines (two 167 MHz UltraSPARC and two
296 MHz SUNW UltraSPARC-II). Using the grammar described here, one
iteration of \texttt{supar} on the entire corpus takes about 2.5
hours, lexicalization and generating an initial lexicalized model
takes more than six hours, and an iteration of lexicalized parsing can
be done in 5.5 hours.


\section{Evaluation}
\label{sec:evaluation}

For the evaluation, a total of 600 randomly selected clauses were
manually annotated by two labellers. Using a chart browser, the
labellers filled the appropriate cells with category names of NCs and
those of maximal VP projections (cf. Figure \ref{fig:muon} for an
example of NC-labelling). Subsequent alignment of the labellers
decisions resulted in a total of 1353 labelled NC categories (with
four different cases). The total of 584 labelled VP categories
subdivides into 21 different verb frames with 340 different lemma
heads. The dominant frames are active transitive (164 occurrences) and
active intransitive (117 occurrences). They represent almost half of
the annotated frames. Thirteen frames occur less than ten times, five
of which just once.

\subsection{Methodology}
\label{sec:eval-methods}

To evaluate iterative training, we extracted maximum probability
(Viterbi) trees for the 600 clause test set in each iteration of
parsing.  For extraction of a maximal probability parse in
unlexicalized training, we used Schmid's \texttt{lopar} parser
\cite{Schmid:99}.  Trees were mapped to a database of parser generated
markup guesses, and we measured \emph{precision} and \emph{recall}
against the manually annotated category names and spans. Precision
gives the ratio of correct guesses over all guesses, and recall the
ratio of correct guesses over the number of phrases identified by
human annotators. Here, we render only the precision/recall results on
pairs of category names and spans, neglecting less interesting
measures on spans alone. For the figures of adjusted recall, the
number of unparsed misses has been subtracted from the number of
possibilities.

In the following, we focus on the combination of the best
unlexicalized model and the lexicalized model that is grounded on the
former.

\subsection{NC Evaluation}
\label{sec:eval-nc}

\begin{figure}[t]
  \hspace{-.8em}\mbox{\psfig{file=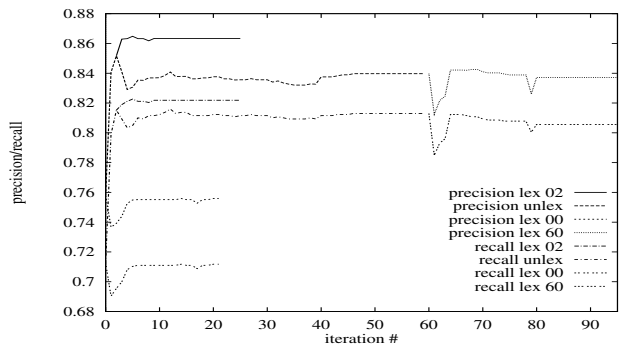,width=8.6cm,height=4.5cm}}

  \begin{center}
    \vspace{-1.5em}
    \caption{Precision/recall measures on NC cases}
    \label{fig:nc-plot}
    \vspace{-1.5em}
  \end{center}
\end{figure}

Figure \ref{fig:nc-plot} plots precision/recall for the training runs
described in section \ref{sec:training}, with lexicalized parsing
starting after 0, 2, or 60 unlexicalized iterations. The best results
are achieved by starting with lexicalized training after two
iterations of unlexicalized training. Of a total of 1353 annotated
\texttt{NC}s with case, 1103 are correctly recognized in the best
unlexicalized model and 1112 in the last lexicalized model. With a
number of 1295 guesses in the unlexicalized and 1288 guesses in the
final lexicalized model, we gain 1.2\% in precision (85.1\% vs.\ 
86.3\%) and 0.6\% in recall (81.5\% vs.\ 82.1\%) through lexicalized
training. Adjustment to parsed clauses yields 88\% vs.\ 89.2\% in
recall. As shown in Figure \ref{fig:nc-plot}, the gain is achieved
already within the first iteration; it is equally distributed between
corrections of category boundaries and labels.

The comparatively small gain with lexicalized training could be viewed
as evidence that the chunking task is too simple for lexical
information to make a difference. However, we find about 7\% revised
guesses from the unlexicalized to the first lexicalized model.
Currently, we do not have a clear picture of the newly introduced
errors.

The plots labeled ``00'' are results for lexicalized training starting
from a random initial grammar. The precision measure of the first
lexicalized model falls below that of the unlexicalized random model
(74\%), only recovering through lexicalized training to equalize the
precision measure of the random model (75.6\%).  This indicates
that some degree of unlexicalized initialization is necessary,
if a good lexicalized model is to be obtained.

\cite{Skut/Brants:98} report 84.4\% recall and 84.2\% for NP and PP
chunking without case labels.  While these are numbers for a simpler
problem and are slightly below ours, they are figures for an
experiment on unrestricted sentences.  A genuine comparison has to
await extension of our model to free text.  

\subsection{Verb Frame Evaluation}
\label{sec:eval-vp}

\begin{figure}[t]
  \hspace{-.8em}\mbox{\psfig{file=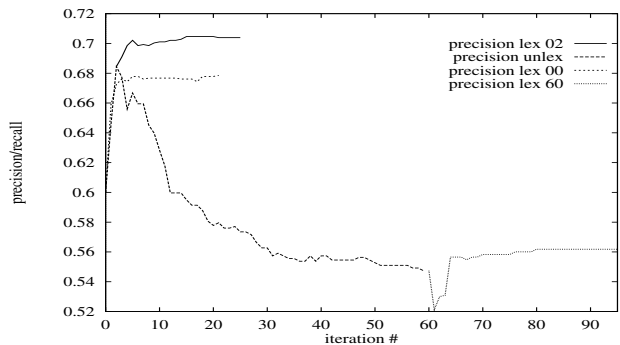,width=8.6cm,height=4.5cm}}

  \begin{center}
    \vspace{-1.5em}
    \caption{Precision measures on all verb frames}
    \label{fig:vp-plot}
    \vspace{-1.5em}
  \end{center}
\end{figure}

Figure \ref{fig:vp-plot} gives results for verb frame recognition under
the same training conditions. Again, we achieve best results by
lexicalising the second unlexicalized model. Of a total of 584
annotated verb frames, 384 are correctly recognized in the best
unlexicalized model and 397 through subsequent lexicalized training.
Precision for the best unlexicalized model is 68.4\%. This is raised
by 2\% to 70.4\% through lexicalized training; recall is 65.7\%/68\%;
adjustment by 41 unparsed misses makes for 70.4\%/72.8\% in recall.
The rather small improvements are in contrast to 88 differences in
parser markup, i.e. 15.7\%, between the unlexicalized and second
lexicalized model. The main gain is observed within the first two
iterations (cf.\ Figure \ref{fig:vp-plot}; for readability, we dropped
the recall curves when more or less parallel to the precision curves).

Results for lexicalized training without prior unlexicalized training
are better than in the NC evaluation, but 
fall short of our best results by more than 2\%.

The most notable observation in verb frame evaluation is the decrease
of precision of frame recognition in unlexicalized training from the
second iteration onward.  After several dozen iterations, results are
5\% below a random model and 14\% below the best model. The primary
reason for the decrease is the mistaken revision of adjoined PPs to
argument PPs. E.g.\ the required number of 164 transitive frames is
missed by 76, while the parser guesses 64 \texttt{VPA.nap} frames in
the final iteration against the annotator's baseline of 12. In
contrast, lexicalized training generally stabilizes w.r.t.\ frame
recognition results after only few iterations.

The plot labeled ``lex 60'' gives precision for a lexicalized training
starting from the unlexicalized model obtained with 60 iterations,
which measured by linguistic criteria is a very poor state.
As far as we know, lexicalized EM estimation never recovers from
this bad state. 

\subsection{Evaluation of non-PP Frames}

Because examination of individual cases showed that PP attachments are
responsible for many errors, we did a separate evaluation of non-PP
frames. We filtered out all frames labelled with a PP argument from
both the maximal probability parses and the manually annotated frames
(91 filtered frames), measuring precision and recall against the
remaining 493 labeller annotated \mbox{non-PP} frames.

\begin{figure}[t]
  \hspace{-.8em}\mbox{\psfig{file=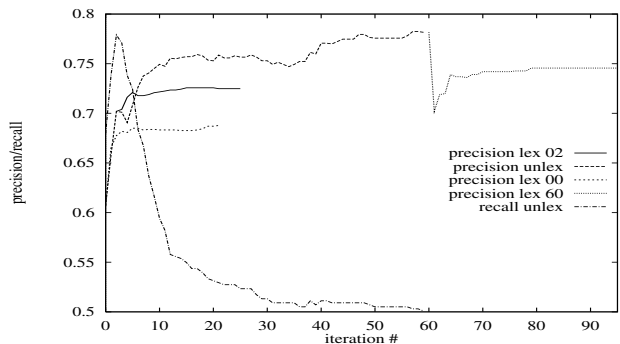,width=8.6cm,height=4.5cm}}
  \begin{center}
    \vspace{-1.5em}
    \caption{Precision measures on non-PP frames}
    \label{fig:no_p-plot}
    \vspace{-1.5em}
  \end{center}
\end{figure}

For the best lexicalized model, we find somewhat but not excessively
better results than those of the evaluation of the entire set of
frames. Of 527 guessed frames in parser markup, 382 are correct, i.e.\ 
a precision of 72.5\%. The recall figure of 77.5\% is considerably
better since overgeneration of 34 guesses is neglected. The
differences with respect to different starting points for
lexicalization emulate those in the evaluation of all frames.

The rather spectacular looking precision and recall differences in
unlexicalized training confirm what was observed for the full frame
set. From the first trained unlexicalized model throughout
unlexicalized training, we find a steady increase in precision (70\%
first trained model to 78\% final model) against a sharp drop in
recall (78\% peek in the second model vs.\ 50\% in the final).
Considering our above remarks on the difficulties of frame recognition
in unlexicalized training, the sharp drop in recall is to be expected:
Since recall measures the correct parser guesses against the
annotator's baseline, the tendency to favor PP arguments over
PP-adjuncts leads to a loss in guesses when PP-frames are abandoned.
Similarly, the rise in precision is mainly explained by the decreasing
number of guesses when cutting out non-PP frames. For further
discussion of what happens with individual frames, we refer the reader
to \cite{Beil:98}.

One systematic result in these plots is that performance of
lexicalized training stabilizes after a few iterations.
This is consistent with what happens with rule
parameters for individual verbs, which are close
to their final values within five iterations.

\section{Conclusion} 
Our principal result is that scrambling-style free-er phrase order,
case morphology and subcategorization, and NP-internal gender, number
and case agreement can be dealt with in a head-lexicalized PFCG
formalism by means of carefully designed categories and rules which
limit the size of the packed parse forest and give desirable pooling
of parameters. Hedging this, we point out that we made compromises in
the grammar (notably, in not enforcing nominative-verb agreement) in
order to control the number of categories, rules, and parameters.

A second result is that iterative lexicalized inside-outside
estimation appears to be beneficial, although the precision/recall
increments are small.  We believe this is the first substantial
investigation of the utility of iterative lexicalized inside-outside
estimation of a lexicalized probabilistic grammar involving a
carefully built grammar where parses can be evaluated by linguistic
criteria.

A third result is that using too many unlexicalized iterations (more
than two) is detrimental. A criterion using cross-entropy
overtraining on held-out data dictates many more unlexicalized
iterations, and this criterion is therefore inappropriate.

Finally, we have clear cases of lexicalized EM estimation being stuck
in linguistically bad states.  As far as we know, the model which gave
the best results could also be stuck in a comparatively bad state.
We plan to experiment with other lexicalized training regimes,
such as ones which alternate between different training corpora.

The experiments are made possible by improvements in parser and
hardware speeds, the carefully built grammar, and evaluation tools. In
combination, these provide a unique environment for investigating
training regimes for lexicalized PCFGs. Much work remains to be done
in this area, and we feel that we are just beginning to develop
understanding of the time course of parameter estimation, and of the
general efficacy of EM estimation of lexicalized PCFGs as evaluated by
linguistic criteria.

We believe our current grammar of German could be extended to a robust
free-text chunk/phrase grammar in the style of the English grammar of
Carroll and Rooth (1998) with about a month's work, and to a free-text
grammar treating verb-second clauses and additional complementation
structures (notably extraposed clausal complements) with about 
one year of additional grammar development and experiment.  These
increments in the grammar could easily double the number of rules.
However this would probably not pose a problem for the parsing and
estimation software.

\nocite{Skut/Brants:98}

\end{document}